\documentclass[lettersize,journal]{IEEEtran}
\usepackage{cite}
\usepackage{amsmath,amsfonts}
\usepackage{array}
\usepackage{ragged2e}
\usepackage{algorithm}
\usepackage{algpseudocode}
\usepackage{graphicx}
\usepackage{tabularx}
\usepackage{textcomp}
\usepackage{url}
\usepackage{booktabs}
\usepackage{color}
\usepackage[hidelinks]{hyperref}

\newcommand{\VarOurs}{\textsc{Ours}}
\newcommand{\VarNoSys}{\VarOurs{} w/o System~2}
\newcommand{\VarSingle}{\VarOurs{} w/o Temporal Context}
\newcommand{\VarNoRec}{\VarOurs{} w/o Replanning}
\newcommand{\VarTimeout}{\VarOurs{} w/o Learned Trigger}
\newcommand{\VarOracle}{\textsc{Oracle Pointer}}
\newcommand{\VarVLM}{\textsc{VLM-Monitor}}

\title{\LARGE \bfseries
Robo-MLT: Dual-System Robotic Manipulation for Long-Horizon Medical Laboratory Tasks
}
\author{%
Zequan Zeng\textsuperscript{1},
Kai Wang\textsuperscript{1,*},
Delvin Ce Zhang\textsuperscript{2},
Huaping Liu\textsuperscript{3},
Zhipeng Yuan\textsuperscript{2},
Junqin Yang\textsuperscript{1}, YunLong Zhang\textsuperscript{1}%
\thanks{\textsuperscript{1}School of Automation, Chongqing University, Chongqing, 400044, China.}%
\thanks{\textsuperscript{2}School of Computer Science, The University of Sheffield, Sheffield, S1 4DP, United Kingdom.}%
\thanks{\textsuperscript{3}Department of Computer Science and Technology, Tsinghua University, Beijing, 100084, China.}%
\thanks{\textsuperscript{*}Corresponding author: Kai Wang.}%
}

\begin{document}
\bstctlcite{IEEEexample:BSTcontrol}
\maketitle
\thispagestyle{empty}

\begin{abstract}
 Reliable medical laboratory automation requires robots to complete entire workflows in protocol order while handling changes during execution. A robot may execute each motion correctly yet fail the workflow by confusing visually similar stages or stalling after human intervention. We present Robo-MLT, a dual-system framework that couples explicit workflow monitoring with learned action generation.
A vision--language planner composes an ordered atomic-skill sequence from the protocol and the initial workspace observation. A multiframe execution monitor detects when the active skill has completed or become infeasible, then advances the workflow or triggers replanning. A skill-conditioned vision--language--action policy generates the manipulation trajectories, and an asynchronous runtime coordinates it with the monitor.
We evaluate Robo-MLT against four policy baselines and a VLM-based monitor baseline on blood gas analysis and blood sample tube sorting, with 50 nominal configurations per method and 20 sample-removal configurations. Robo-MLT achieves task success rates of 82\% and 70\%, exceeding $\pi_{0.5}$ by 30 and 22 percentage points, respectively. After an operator removes the sample being handled, the system revises the remaining plan and continues processing the other samples. These results indicate that, under the evaluated conditions, explicit workflow monitoring makes long-horizon laboratory manipulation more reliable and recoverable. Project page: \url{https://robomlt.github.io/Robo-MLT/}.
\end{abstract}

{\renewcommand{\abstractname}{Note to Practitioners}
\begin{abstract}
This work is motivated by the many short, strictly ordered sample-handling procedures that clinical laboratories still perform by hand. Each procedure is simple on its own, but it must be carried out in a fixed order, and a step performed out of order can invalidate a result. Dedicated automation handles high and stable sample volumes well, yet every new protocol, consumable, or bench layout has to be engineered into it. This paper describes a robot that instead keeps the protocol as an explicit, ordered list of named operations. A planner reads the protocol and the initial view of the bench and writes the list; a monitor watches the camera streams and decides when the current operation has finished, or when something in the workspace has made it impossible to finish; a learned controller performs each operation. Because the planner is not retrained, a laboratory can reorder or repeat operations that the robot already knows---for example, run the same assay for however many tubes happen to be on the rack---without collecting data or retraining anything. Teaching a genuinely new operation is more expensive: it requires teleoperated demonstrations and retraining of the monitor. The system requires a dual-arm manipulator, three RGB cameras, and a single workstation-class GPU. Its main demonstrated practical benefit is workflow continuity: when an operator removes a sample from the robot's gripper during execution, the system can revise the remaining plan and continue processing other samples rather than becoming stalled. Current validation is limited to two laboratory protocols, a single workspace, and one type of external intervention, without deployment in a production laboratory. Future work will focus on expanding the atomic-skill library and evaluating the system across a wider range of disturbances and laboratory configurations.
\end{abstract}}

\section{Introduction}
\label{sec:intro}
Medical laboratory automation aims to reduce repetitive manual work, increase testing throughput, and limit personnel exposure to clinical samples. These benefits require robots to complete entire workflows reliably, not merely to demonstrate isolated manipulation skills. In blood gas analysis, for example, a robot must handle a sample, prepare it for testing, interact precisely with the analyzer, and then withdraw and discard it in the prescribed order. These operations depend on one another: a successful grasp does not ensure a successful test, and a physically feasible action may still be wrong for the current workflow stage. Reliable automation therefore requires both accurate manipulation and consistent execution of the operating protocol~\cite{zhang2025robochemist,dai2024autonomous,boiko2023autonomous}.

\begin{figure}[!t]
    \centering
    \includegraphics[width=\columnwidth]{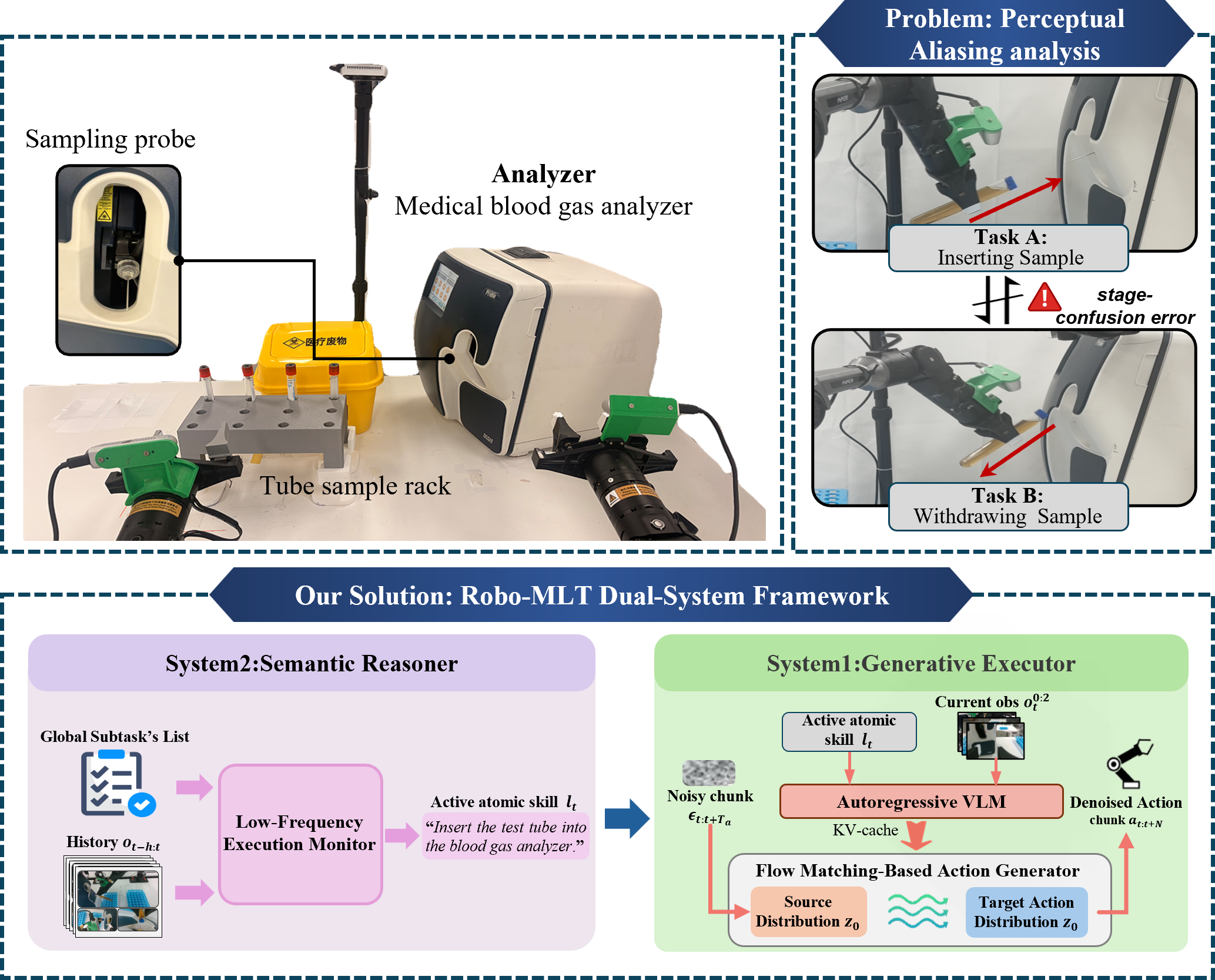}
    \caption{\textbf{Perceptual aliasing in medical laboratory automation, and the proposed Robo-MLT framework.}
        (Top) In a reversible workflow, visually similar observations can require opposite actions; a policy without explicit workflow state may therefore act at the wrong stage. (Bottom) System~2 tracks the active skill from temporal observations; System~1 maps that skill and the current observation to fine-grained trajectories.}
    \label{fig:intro_figure}
\end{figure}

Meeting these requirements is difficult in a laboratory workspace shared with people. Sample positions vary, instrument interfaces demand precise alignment, and operators may intervene while a task is running. Different workflow stages can also produce nearly identical observations: a sample at an analyzer inlet must be inserted before testing but withdrawn afterward, yet the local images look alike (Fig.~\ref{fig:intro_figure}). Choosing the correct action therefore requires knowing what has already been completed. Disturbances add a further difficulty. A displaced target may only require the ongoing motion to be adjusted, whereas removing the sample from the gripper makes the current step impossible to complete and forces the remaining workflow to be revised.

Existing approaches relevant to these requirements fall into three categories. The first is fixed-program and task-specific laboratory automation, which implements procedures through predefined control logic, calibrated motions, dedicated fixtures, and purpose-built mechanisms~\cite{burger2020mobile,thieme2024deep}. In clinical laboratories, total laboratory automation lines handle tube transport and sample preparation while human operators oversee the workflow and make non-routine decisions~\cite{brown2022next}. The same principle underlies task-specific devices such as microtube cappers~\cite{jinno2024manually}, the KANKA tube sorter~\cite{sisman2024kanka}, and swab-sampling robots~\cite{sun2023autonomous}, while research architectures encode the workflow as state machines~\cite{fakhruldeen2022archemist}, behavior trees~\cite{fakhruldeen2025multimodal}, or modular integration frameworks~\cite{wolf2024towards}. Within their intended operating conditions, these designs provide procedural consistency and predictable cycle times by construction; because every transition is enumerated in advance, however, a new workspace configuration or an unanticipated contingency requires the sensing, control logic, and recovery branches to be authored by hand for each task. Demonstration-trained manipulation skills offer a complementary route: they generate actions from visual observations and can be reused across supported variations in object pose without reprogramming individual trajectories~\cite{takahashi2025ai}.

The second category comprises end-to-end imitation-learning and vision--language--action (VLA) policies. Methods such as ACT~\cite{zhao2023learning}, Diffusion Policy~\cite{chi2025diffusion}, $\pi_0$~\cite{black2024pi_0}, and SmolVLA~\cite{shukor2025smolvla} learn actions from demonstrations, and VLA models add language conditioning and pretrained visual--semantic representations. Because they act on the current observation, they suit the manipulation studied here: tube poses differ between runs, and seating a tube in an analyzer inlet with roughly 5\,mm of clearance requires alignment from the live view rather than a calibrated waypoint. Their language conditioning also offers a natural interface to a higher-level component. A policy that does not explicitly maintain workflow progress, however, must infer the current stage from its observations and conditioning inputs. When stages look alike, insufficient temporal context can lead to repeated or stage-inconsistent actions. Moreover, adapting a trajectory to a displaced target does not determine how the workflow should proceed once the current operation is no longer feasible.

The third category separates high-level planning or semantic reasoning from low-level execution. In laboratory robotics, ORGANA~\cite{darvish2025organa} combines language-guided planning with visual feedback, and RoboChemist~\cite{zhang2025robochemist} couples vision--language model (VLM) planning and monitoring with VLA control, advancing once a subtask is verified and repeating it otherwise. General hierarchical systems such as RoboDual~\cite{bu2024robodual}, Fast-in-Slow~\cite{chen2025fast}, and $\pi_{0.5}$~\cite{intelligence2025pi_05} connect semantic decisions to continuous manipulation. Recent VLA frameworks also bring progress and recovery inside the policy: PALM~\cite{liu2026palm} predicts within-subtask progress jointly with actions, and TCoT~\cite{li2026tcot} couples success assessment with trajectory-level recovery. Progress representations and recovery mechanisms therefore already exist; systems differ in the level at which they act. For a prescribed laboratory protocol, the governing decisions are discrete and ordered: which step is active, whether it is finished, and which of the remaining steps should follow when an intervention makes the active step impossible.

Taken together, these lines of work leave three connected requirements for reliable medical laboratory manipulation. The first is \emph{workflow-stage identification}: the robot must distinguish visually similar stages using execution history and protocol progress, so that finishing one step leads to the next step rather than to a repetition. The second is \emph{recovery within the protocol}: when a disruption makes the active skill impossible to complete, the robot must recognize this promptly and revise the remaining steps in the order the protocol prescribes. The third is \emph{coordination of monitoring and manipulation}: workflow decisions rest on temporal observations and semantic assessment, whereas physical execution needs an uninterrupted action stream, so the two must run at different rates without the slower one stalling the arm.

To this end, we propose \emph{Robo-MLT} (Robotic Medical Laboratory Tasks), a hierarchical dual-system framework inspired by dual-process theory~\cite{kahneman2011thinking} and summarized in Fig.~\ref{fig:intro_figure} (bottom). 
A frozen vision--language planner first composes an ordered plan over an atomic-skill library from the task, the operating protocol, and the initial workspace observation. System~2 tracks the active plan step with a lightweight multiframe monitor. Its \emph{completion head} confirms that the active skill has finished, while its \emph{recovery head} is trained to detect states in which that skill can no longer be completed; a stall timeout serves as a backup trigger. Recovery then replans the remaining steps from the protocol, the completed steps, and the current observation. System~1 generates continuous action chunks conditioned on the active skill and the current observations, and an asynchronous runtime streams them at the robot control rate.

\begin{figure*}[!t]
    \centering
    \includegraphics[width=\textwidth]{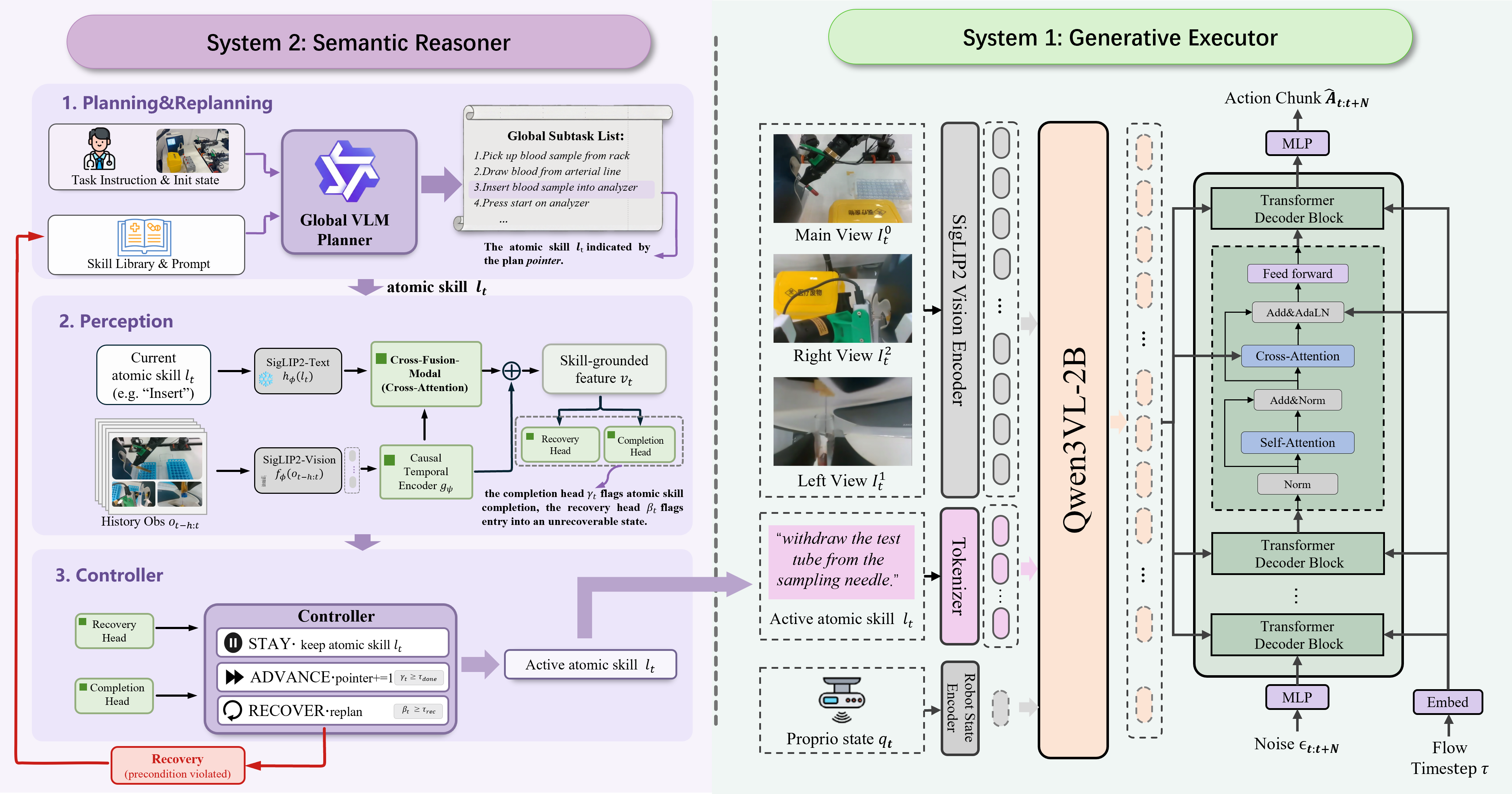}
    \caption{\textbf{Overview of Robo-MLT.} A frozen Qwen3.7-Max planner maps a user command, the instrument protocol, and an initial scene observation to an ordered sequence of atomic skills. System~2 tracks the plan with a dual-head execution monitor and invokes replanning once the active skill can no longer be completed; System~1 executes the active skill with a Qwen3-VL-conditioned flow-matching policy.}
    \label{fig:framework_overview}
\end{figure*}
The main contributions are as follows:
\begin{itemize}
   \item We propose Robo-MLT, a hierarchical framework that separates explicit, low-frequency workflow monitoring from high-frequency generative control for long-horizon medical laboratory manipulation.

    \item We develop a lightweight multiframe execution monitor whose two decision heads determine when the active skill has completed and when it can no longer be completed, in which case the remaining workflow is replanned from the protocol and the current scene.

    \item We evaluate Robo-MLT on two real-world tasks, blood gas analysis and blood sample tube sorting, through comparative and ablation experiments, an offline switch-timing analysis of the monitor, and sample-removal disturbance tests, reporting end-to-end task success and recovery success after intervention.
\end{itemize}

\section{Related Work}
\subsection{Laboratory Automation and Medical Robotics}
Laboratory automation seeks to improve throughput, reproducibility, and personnel safety, with recent robotic systems extending automation to manipulation and coordination across laboratory instruments. Dai et al.~\cite{dai2024autonomous} integrated mobile robots with synthesis and analytical equipment for autonomous exploratory chemistry. At the manipulation level, Schober et al.~\cite{schober2025vision} combined visual liquid-volume estimation with simulation-driven pouring and demonstrated their integration into robotic cell-culture automation. These studies illustrate complementary advances in laboratory orchestration and perception-guided physical interaction.
Language-guided systems further connect experimental goals to executable operations. ORGANA~\cite{darvish2025organa} combines interaction with chemists, task planning, and visual feedback for chemistry experimentation, while TARMAC~\cite{huang2025tarmac} organizes laboratory manipulations into reusable primitives and higher-level macros. RoboChemist~\cite{zhang2025robochemist} integrates VLM-based task decomposition, visual prompting, and completion and compliance assessment with VLA control. Its execution loop advances after successful subtasks and repeats unsuccessful ones, demonstrating that feedback and corrective execution are already important components of robotic laboratory automation.
In medical laboratories, KANKA~\cite{sisman2024kanka} addresses blood-tube sorting and preanalytical quality control, providing a directly relevant example of task-specific clinical laboratory robotics. AutoBio~\cite{lan2025autobio} provides a simulation benchmark for protocol-based manipulation in biology laboratories, while RoboNurse-VLA~\cite{li2025robonurse} explores VLA-based surgical instrument handover in an adjacent medical setting.

Building on these advances, Robo-MLT studies the execution of blood gas analysis and sample-tube sorting with an explicit protocol-grounded skill sequence. Its emphasis is on maintaining the active protocol step, verifying completion from temporal observations, and revising the remaining sequence when an intervention disrupts execution.

\subsection{Vision--Language--Action Models and Hierarchical Manipulation}
Learning-based robotic manipulation has progressed from task-specific imitation learning to general-purpose vision--language--action (VLA) models. ACT~\cite{zhao2023learning} employs action chunking to improve fine-grained bimanual manipulation, while Diffusion Policy~\cite{chi2025diffusion} models multimodal robot actions through iterative denoising. Building on pretrained vision--language representations, RT-2~\cite{zitkovich2023rt} and OpenVLA~\cite{kim2024openvla} generate robotic actions using autoregressive transformers. More recent models, including $\pi_0$~\cite{black2024pi_0} and SmolVLA~\cite{shukor2025smolvla}, adopt diffusion- or flow-based action generation to produce continuous trajectories. Although these approaches improve manipulation flexibility and instruction following, a policy without an explicit workflow-state representation must infer the active stage from appearance and conditioning inputs alone, which is difficult when different stages of a long-horizon workflow look alike.

Hierarchical and dual-system architectures coordinate semantic reasoning with low-level control through intermediate representations ranging from latent features to language-based subtask commands. RoboDual~\cite{bu2024robodual} conditions a specialist diffusion policy on a generalist VLA's latent representations and discretized actions, while GR00T N1~\cite{nvidia2025groot} jointly trains a vision--language module and a diffusion-transformer action module. Fast-in-Slow~\cite{chen2025fast} integrates fast execution within a slower VLM through partial parameter sharing and asynchronous inference. At the language level, Hi Robot~\cite{shi2025hirobot} translates complex instructions and situated user feedback into commands for a low-level VLA, whereas $\pi_{0.5}$~\cite{intelligence2025pi_05} uses a shared model to predict semantic subtasks and generate action chunks conditioned on those predictions.

While these architectures improve the coordination between semantic reasoning and motor control, explicit workflow-state tracking remains a challenge. Robo-MLT addresses this gap by combining protocol-grounded skill planning with temporal monitoring for step transitions and recovery, while conditioning a generative action policy on the active skill instruction.

\section{Methodology}
Determining whether to advance or revise the current protocol step requires discrete decisions based on temporal observation history, whereas executing the active step requires a continuous stream of actions at the robot control rate. Inspired by the fast--slow separation of dual-process theory, Robo-MLT assigns these responsibilities to two complementary systems. As shown in Fig.~\ref{fig:framework_overview}, \textbf{System~2} is a low-frequency execution monitor that tracks the plan and maintains the active atomic skill, and \textbf{System~1} is a high-frequency generative executor that maps the active skill to continuous actions. A top-level planner initializes the ordered skill sequence, while an asynchronous runtime coordinates the two systems.
\begin{figure*}[!t]
    \centering
    \includegraphics[width=\textwidth]{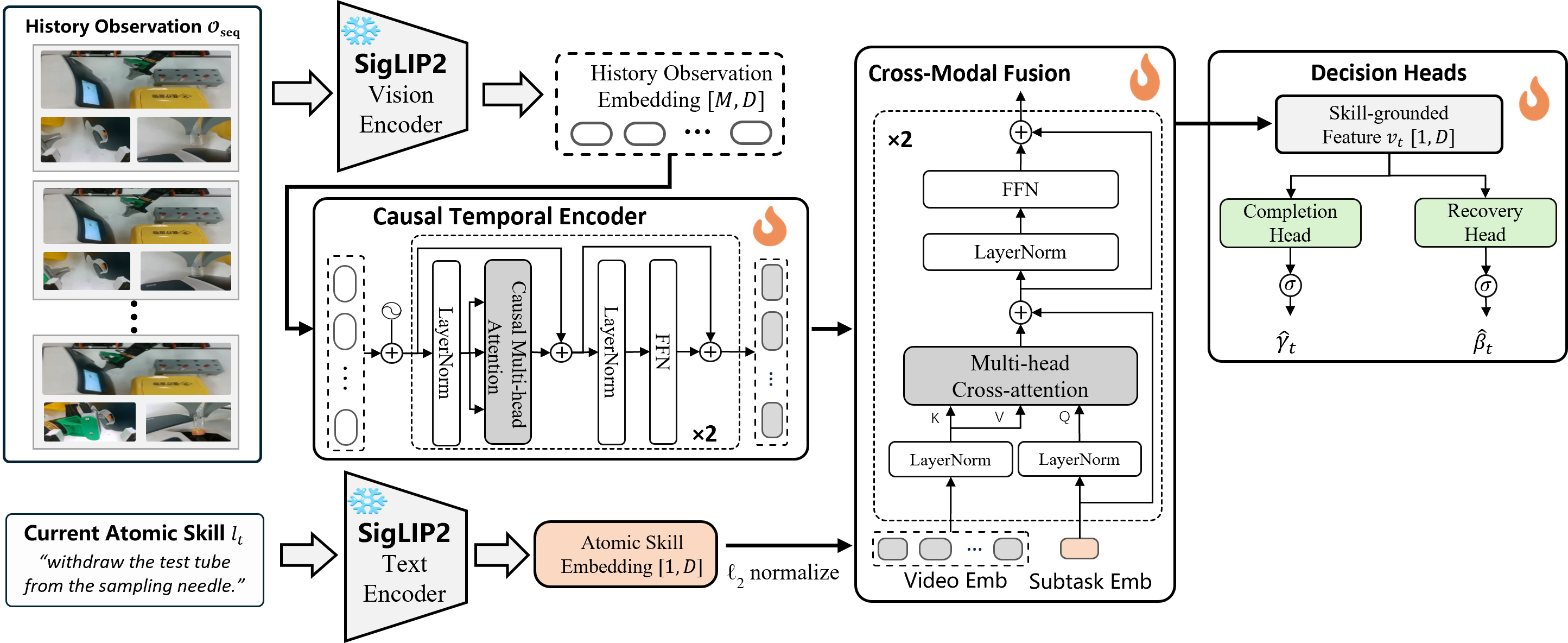}
    \caption{\textbf{Architecture of the System~2 execution monitor.} Frozen SigLIP2 encoders embed a sliding window of $M$ multiview observations and the active skill instruction $l_t$; a causal temporal encoder and cross-attention blocks ground the observation history in the instruction to form the feature $v_t$, from which two MLP heads predict the completion and recovery probabilities $\hat{\gamma}_t$ and $\hat{\beta}_t$.}
    \label{fig:monitor_model}
\end{figure*}
\subsection{Top-Level Task Planning over an Atomic-Skill Library}
Rather than issuing unrestricted free-form commands during execution, Robo-MLT grounds high-level planning in a declarative atomic-skill library:
\begin{equation}
    \mathcal{S}=\{s_1,\ldots,s_{|\mathcal{S}|}\}
\end{equation}
where $|\mathcal{S}|$ is the number of atomic skills in the library and each skill $s_j$ is specified by a canonical natural-language instruction $l_j$; here $|\mathcal{S}|=7$ for blood gas analysis and $|\mathcal{S}|=5$ for tube sorting. Given a task instruction, the corresponding instrument protocol, and an initial observation $I_0$ of the workspace, a frozen vision--language planner $\Phi_{\mathrm{plan}}$ composes an ordered skill sequence over the library:
\begin{equation}
    \label{equ:plan}
    \mathcal{P}=\Phi_{\mathrm{plan}}(\text{task},\,\text{protocol},\,\mathcal{S},\,I_0)
    =\big(s_{\kappa(1)},\,\ldots,\,s_{\kappa(|\mathcal{P}|)}\big)
\end{equation}
where $|\mathcal{P}|$ is the plan length and $\kappa(i)$ maps the $i$-th step to a skill in $\mathcal{S}$; $\kappa$ need not be injective, since a skill may recur at several steps of the same plan. We instantiate $\Phi_{\mathrm{plan}}$ with the Qwen3.7-Max API~\cite{yang2025qwen3}, queried zero-shot and never fine-tuned; the same model serves as the replanner $\Phi_{\mathrm{replan}}$ below.
Conditioning on $I_0$ lets the planner instantiate the plan from the observed scene rather than a fixed layout: it reads the number, positions, and categories of the tubes on the rack and expands the sequence accordingly---repeating the assay for each detected tube in blood gas analysis, and committing to a feasible pick-and-place order in tube sorting. Because the planner is frozen and System~2 only moves the pointer, any re-ordering or repetition of the skills in $\mathcal{S}$ yields a workflow that the same trained components execute without a gradient update.

\subsection{System 2: Low-Frequency Execution Monitor}
System~2 monitors the execution of the ordered plan $\mathcal{P}$. At each update, it reads a short observation history and the instruction of the active skill and returns one of $\{\textsc{stay},\textsc{advance},\textsc{recover}\}$: it holds the plan pointer $\mathrm{ptr}$, advances it after the skill completes, or invokes replanning once the skill can no longer be completed. The pointer fixes the active skill, and the observation history decides when the pointer moves. Visually similar states that require different actions, such as tube insertion and withdrawal, are therefore distinguished by their position in the workflow rather than by appearance alone. The monitor consists of a temporal skill-grounding trunk followed by two decision heads (Fig.~\ref{fig:monitor_model}); a \textsc{recover} decision additionally invokes the VLM replanner described below.

\subsubsection{Temporal Skill-Grounding Trunk}

The trunk relates a short observation history to the active skill text. Let $\mathcal{O}_{\mathrm{seq}}=\{I_1,\ldots,I_M\}$ denote the $M$ multiview observations sampled by the asynchronous runtime (Sec.~\ref{subsec:async}). A frozen SigLIP2-300M image encoder~\cite{tschannen2025siglip2} $f_\phi(\cdot)$ embeds each sampled observation, and a Transformer trunk~\cite{vaswani2017attention} $g_\psi(\cdot)$ aggregates the resulting sequence while conditioning on the active instruction. As shown in Fig.~\ref{fig:monitor_model}, $g_\psi$ operates in two stages: a causal temporal encoder with learned positional encodings mixes the observation features along time, after which cross-attention blocks ground that history in the instruction, taking the text embedding as the query and the temporal features as keys and values. The trunk therefore returns a single skill-grounded vector rather than a pooled summary of the window:
\begin{equation}
    v_t = g_{\psi}\!\left(
        \left[\, f_{\phi}(I)\, \right]_{I \in \mathcal{O}_{\mathrm{seq}}},\;
        h_{\phi}(l_t)
    \right)
\end{equation}
where $h_{\phi}(\cdot)$ is the frozen SigLIP2 text encoder whose output is $\ell_2$-normalized, and $l_t$ is the canonical instruction associated with the active skill $s_{\kappa(\mathrm{ptr})}$.

\subsubsection{Decision Heads, Recovery, and Training}
\label{subsec:heads}
Two lightweight binary classification heads operate on the shared representation $v_t$. The completion head $\hat{\gamma}_t$ estimates whether the active skill has finished, whereas the recovery head $\hat{\beta}_t$ detects states in which continuing the active skill is unlikely to succeed, whether because an operator has disrupted the scene or because execution itself has failed:
\begin{equation}
    \hat{\gamma}_t = \sigma\!\left(\psi_{\mathrm{comp}}(v_t)\right), \qquad
    \hat{\beta}_t = \sigma\!\left(\psi_{\mathrm{rec}}(v_t)\right)
\end{equation}
where $\hat{\gamma}_t,\hat{\beta}_t\in[0,1]$, and $\psi_{\mathrm{comp}}$ and $\psi_{\mathrm{rec}}$ are two-layer MLPs that share no parameters. The monitor debounces these probabilities: it returns \textsc{advance} when $\hat{\gamma}_t \ge \tau_{\mathrm{done}}$ holds for $k_a$ consecutive updates, \textsc{recover} when $\hat{\beta}_t \ge \tau_{\mathrm{rec}}$ holds for $k_b$ consecutive updates, and \textsc{stay} otherwise.

If both conditions fire simultaneously, \textsc{recover} takes priority: the arm holds its position in a replanning state instead of proceeding to the next stage. A stall is handled the same way: once the pointer has rested on one skill for longer than $T_{\max}$, the monitor returns \textsc{recover} regardless of $\hat{\beta}_t$, so the system cannot remain indefinitely on a skill that the completion head never confirms as finished.

When the recovery head fires or the active skill times out, that skill is treated as no longer completable (e.g., after an operator removes the in-hand tube, or after a slipped grasp leaves the gripper empty). The monitor then invokes the VLM replanner $\Phi_{\mathrm{replan}}$, which revises the remaining plan from the completed plan prefix $\mathcal{C}_t$, a short-term record $\mathcal{R}_t$ of attempted skills, and the current observation $I_t$:
\begin{equation}
    \label{equ:replan}
    \mathcal{P}' = \Phi_{\mathrm{replan}}\big(\text{task},\,\text{protocol},\,\mathcal{S},\,\mathcal{C}_t,\,\mathcal{R}_t,\,I_t\big)
\end{equation}
Conditioning on $I_t$ lets the revised plan refer to the scene as it now stands rather than as it was at $I_0$. The revised steps replace the remainder of the plan, $\mathcal{P}\leftarrow\mathcal{C}_t\oplus\mathcal{P}'$, where $\oplus$ denotes concatenation; because $|\mathcal{C}_t|=\mathrm{ptr}-1$, the pointer already indexes the first revised step and needs no reset.

The skill-grounding trunk $g_\psi$ and the two heads are trained jointly in a single stage using binary cross-entropy, while the SigLIP2 image and text encoders remain frozen. The System~2 objective is:
\begin{equation}
    \label{equ:eq6}
    \begin{aligned}
        \mathcal{L}_{\mathrm{S2}} = -\,\mathbb{E}_{t}\Big[\,
        & \lambda_c\big( w_c\,\gamma_t \log \hat{\gamma}_t + (1-\gamma_t)\log(1-\hat{\gamma}_t)\big) \\
        +\;& \lambda_r\big( w_r\,\beta_t \log \hat{\beta}_t + (1-\beta_t)\log(1-\hat{\beta}_t)\big)\,\Big]
    \end{aligned}
\end{equation}
where $\gamma_t,\beta_t\!\in\!\{0,1\}$ are the labels of the two heads. The coefficients $\lambda_c$ and $\lambda_r$ balance the two heads, while $w_c,w_r\!>\!1$ up-weight the less frequent positive frames. The completion label $\gamma_t$ marks the final frames of a successfully completed segment, while positive labels for $\beta_t$ are drawn from the separately recorded disturbed and failed segments described in Sec.~\ref{subsec:data}.

\subsection{System 1: Generative Executor via Flow Matching}
System~1 maps the atomic instruction selected by System~2 to a continuous trajectory. Following the VLM-conditioned generation paradigm~\cite{black2024pi_0}, it couples a pretrained Qwen3-VL-2B backbone~\cite{qwen2025qwen3vl} with a flow-matching action head.

\subsubsection{VLM Backbone and Semantic Feature Extraction}
At each policy update, Qwen3-VL processes the multiview observation $\mathcal{I}_t=\{I_t^0,I_t^1,I_t^2\}$ and the current instruction $l_t$, producing a semantic feature sequence $h_t=f_{\mathrm{VLM}}(\mathcal{I}_t,l_t)\in\mathbb{R}^{L\times D}$. The robot proprioceptive state $q_t\in\mathbb{R}^{D_q}$ is projected by a multilayer perceptron, $z_q=\mathrm{MLP}(q_t)\in\mathbb{R}^{D}$, and concatenated with $h_t$ to form the conditioning sequence
\begin{equation}
    c_t = [h_t;\, z_q] \in \mathbb{R}^{(L+1)\times D}
\end{equation}
At deployment, $q_t$ is supplied by the commanded joint target $\bar{a}$, the last action in the queue (or the last executed action $a_{\mathrm{prev}}$ when the queue has been cleared), i.e., the target the arm is expected to have reached when the new chunk takes over (Algorithm~\ref{alg:rmlt_dual_system}).

\subsubsection{Action Generation via Flow Matching}
We model the conditional action distribution $p(A\mid c_t)$ with flow matching, where $A\in\mathbb{R}^{N\times D_a}$ is an action chunk, $N$ is the prediction horizon, and $D_a$ is the action dimension. A linear conditional path interpolates between Gaussian noise $\epsilon\sim\mathcal{N}(0,I)$ and the ground-truth action sequence $A_{\mathrm{gt}}$:
\begin{equation}
    A_\tau = \tau A_{\mathrm{gt}} + (1-\tau)\epsilon, \quad \tau \in [0,1]
\end{equation}
The action head $u_\theta$ predicts the corresponding velocity field $u=A_{\mathrm{gt}}-\epsilon$ and is trained by minimizing
\begin{equation}
    \mathcal{L}_{S1}
    = \mathbb{E}_{\tau \sim \mathcal{U}[0,1],\, A_{\mathrm{gt}},\, \epsilon}\!\left[
      \|u_\theta(A_\tau,\tau,c_t) - u\|^2
    \right]
\end{equation}
At inference, numerical integration from $\tau=0$ to $\tau=1$ transforms the initial noise into the predicted action chunk $\hat{A}_{t:t+N}$.

\begin{algorithm}[!t]
\caption{Dual-System Asynchronous Inference}
\label{alg:rmlt_dual_system}
\algtext*{EndIf}\algtext*{EndLoop}
\begin{algorithmic}[1]
    \State \textbf{Init:} $\mathrm{ptr}\leftarrow 1$ on $\mathcal{P}$ (Eq.~\eqref{equ:plan}), $l_{\mathrm{curr}}\leftarrow l_{\kappa(1)}$, $a_{\mathrm{prev}}\leftarrow q_0$; $\mathcal{A},\mathcal{H},J_1,J_2\leftarrow\emptyset$; $k,t,c\leftarrow 0$
    \Loop
        \State $o_t \leftarrow \textsc{CaptureObs}()$;\ \ $t \leftarrow t+1$;\ \ $c \leftarrow \max(c-1,0)$
        \State $d \leftarrow \textsc{Decide}(J_2)$ \Comment{\textsc{stay} while $J_2$ runs}
        \If{$d \neq \textsc{stay}$} \Comment{switch skill}
            \State \textbf{if} $d = \textsc{recover}$ \textbf{then} $\mathcal{P} \leftarrow \mathcal{C}_t \oplus \mathcal{P}'$ \Comment{Eq.~\eqref{equ:replan}}
            \State \textbf{if} $d = \textsc{advance}$ \textbf{then} $\mathrm{ptr} \leftarrow \mathrm{ptr}+1$
            \State \textbf{if} $\mathrm{ptr} > |\mathcal{P}|$ \textbf{then return} \textsc{done}
            \State $l_{\mathrm{curr}} \leftarrow l_{\kappa(\mathrm{ptr})}$;\ \ $\mathcal{A}, J_1 \leftarrow \emptyset$;\ \ $k \leftarrow 0$;\ \ $c \leftarrow \Delta_{\mathrm{cd}}$
        \EndIf
        \If{$t \bmod K = 0$} \Comment{fixed cadence}
            \State append $o_t$ to $\mathcal{H}$, keep last $M$
            \State \textbf{if} $|\mathcal{H}| = M \wedge c = 0$ \textbf{then} $J_2 \leftarrow \textsc{System2}(\mathcal{H}, l_{\mathrm{curr}})$
        \EndIf
        \If{$J_1 = \emptyset$ \textbf{and} $|\mathcal{A}| - k \le r$} \Comment{prefetch}
            \State $J_1 \leftarrow \textsc{System1}(o_t, \bar{a}, l_{\mathrm{curr}})$
        \EndIf
        \If{$J_1$ has finished \textbf{or} $k = |\mathcal{A}|$} \Comment{splice}
            \State $\mathcal{A} \leftarrow \textsc{Blend}(\textsc{Result}(J_1), a_{\mathrm{prev}}, w)$;\ \ $k \leftarrow 0$
        \EndIf
        \State $a_{\mathrm{prev}} \leftarrow a_k$;\ \ \textsc{Execute}$(a_{\mathrm{prev}})$;\ \ $k \leftarrow k+1$
    \EndLoop
\end{algorithmic}
\end{algorithm}

\subsection{Asynchronous Inference Framework}
\label{subsec:async}
System~2 aggregates observations over a relatively long interval, whereas System~1 must supply actions at the robot control rate. We coordinate these components with the asynchronous runtime in Algorithm~\ref{alg:rmlt_dual_system}; Table~\ref{tab:hyperparams} lists its principal hyperparameters. Each system normally runs as a nonblocking background worker with at most one in-flight job, $J_2$ or $J_1$, and the control loop consumes each result when it becomes available. In Algorithm~\ref{alg:rmlt_dual_system}, assigning a call to $J_i$ launches it in the background, and $\textsc{Result}(J_i)$ returns its output, waiting for it if necessary, and frees the slot; while the loop waits, the arm holds $a_{\mathrm{prev}}$. The shared runtime state consists of the plan pointer $\mathrm{ptr}$, an action queue $\mathcal{A}$ refilled $N$ at a time, the queue index $k$, a history buffer $\mathcal{H}$ of capacity $M$, and the active instruction $l_{\mathrm{curr}}$.

\subsubsection{Low-Frequency Plan Monitoring}
The main loop advances the tick $t$ at every control step and appends a sampled multiview observation to $\mathcal{H}$ every $K$ steps, retaining the most recent $M$, so the sampling cadence does not depend on whether a System~2 job is in flight. Once the buffer is full, each append dispatches an atomic snapshot $\mathcal{O}_{\mathrm{seq}}$ of $\mathcal{H}$, together with $l_{\mathrm{curr}}$, to $\textsc{System2}$ as job $J_2$, so the monitor re-evaluates a sliding window every $K$ steps. $\textsc{Decide}(J_2)$ collects a finished job, applies the debouncing rule and stall timeout of Sec.~\ref{subsec:heads} to its $(\hat{\gamma}_t,\hat{\beta}_t)$, and returns \textsc{stay} while the job is still running; its consecutive-hit counters and the per-skill timer restart whenever the active skill changes. Because a monitor job finishes well within one sampling interval (Sec.~\ref{subsec:runtime}), $J_2$ is always free when a new window is due. An \textsc{advance} decision moves the pointer to the next skill, whereas \textsc{recover} holds the arm and invokes replanning; both decisions update $l_{\mathrm{curr}}$. Whenever $d\neq\textsc{stay}$, the main loop discards the current action chunk and pending System~1 job, and launches no System~2 job for a cooldown interval of $\Delta_{\mathrm{cd}}$ steps to prevent rapid switching. The history buffer is not cleared at a switch: training windows are cut from continuous demonstrations, so a window near the start of a skill likewise contains the end of the preceding one.

\subsubsection{High-Frequency Action Streaming}
System~1 generates $N$ actions per chunk. To hide inference latency, the controller prefetches the next chunk before the current one expires: once $|\mathcal{A}|-k\le r$ it launches a policy request as job $J_1$, while every other cycle simply executes the next queued action. The controller splices in the prefetched chunk as soon as it returns. If the queue empties first, the arm holds $a_{\mathrm{prev}}$ until the pending job returns; a synchronous $\textsc{System1}$ call is issued only when no prefetch is pending, as after a skill switch.

Since consecutive flow-matching chunks are generated independently, switching between them may introduce discontinuities at chunk boundaries. To mitigate this, we blend the first $w$ actions of each incoming chunk with $a_{\mathrm{prev}}$ using monotonically increasing weights, while discarding any unexecuted actions from the previous chunk. The remaining actions are executed without modification, avoiding the persistent lag associated with filtering the entire action sequence.

\section{Experiments}
\subsection{Experimental Setup}

\subsubsection{Hardware Platform}
All real-world experiments use an Agilex dual-arm Static-Aloha platform. Three Intel RealSense D435i cameras provide synchronized RGB observations: one fixed global view of the workspace and two wrist-mounted views for fine-grained manipulation. Both systems are trained on an NVIDIA A100 and deployed on an RTX~4090. All methods share the same camera streams, robot state, action space, controller frequency, and safety limits.

\subsubsection{Tasks and Evaluation Conditions}
\begin{figure}[!t]
    \centering
    \includegraphics[width=\columnwidth]{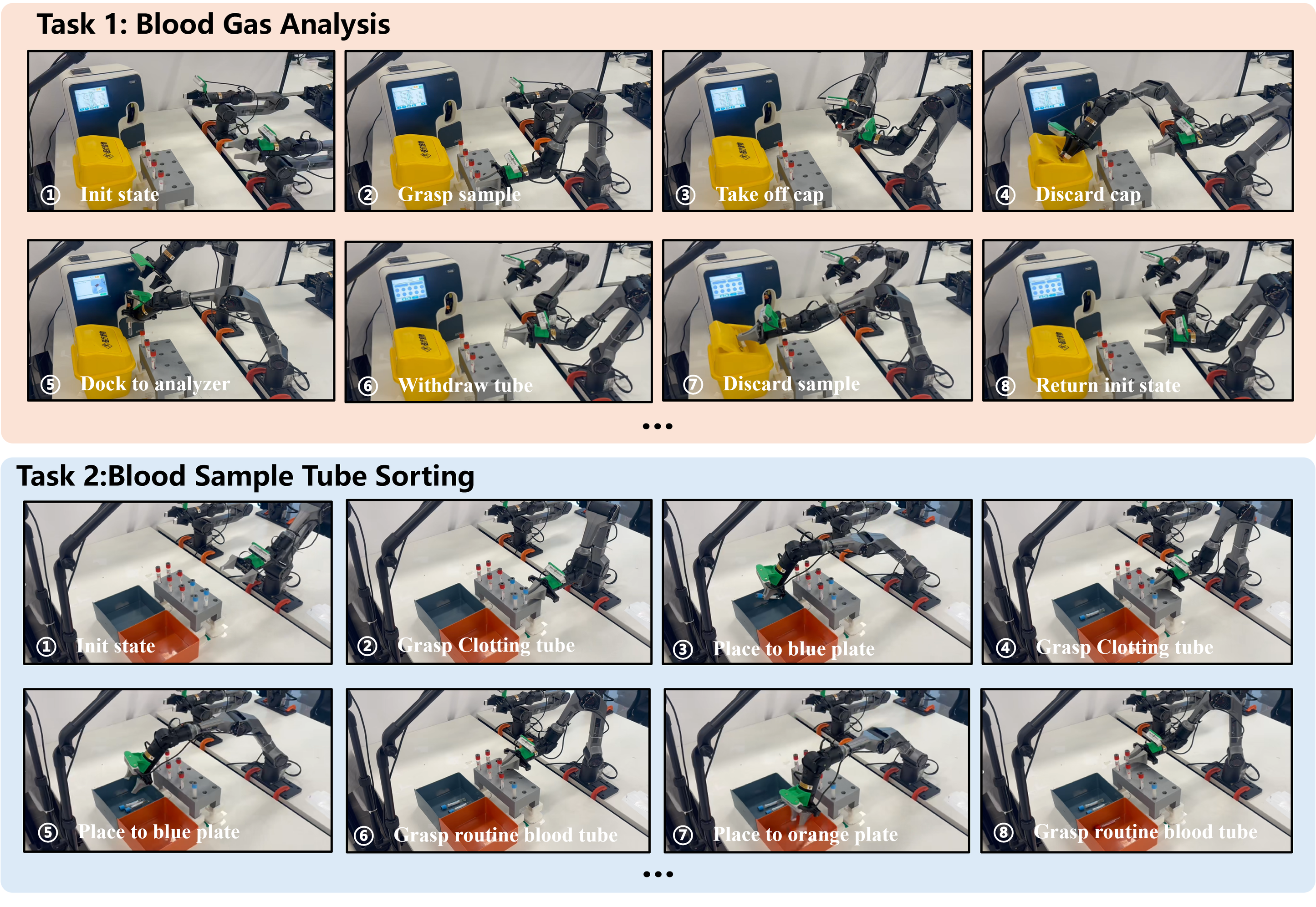}
    \caption{\textbf{The two real-world medical laboratory tasks.}
    \textit{(Top) Blood Gas Analysis} with its $|\mathcal{S}|\!=\!7$ atomic skills.
    \textit{(Bottom) Blood Sample Tube Sorting}: clotting and routine tubes are randomly placed on the rack and sorted into their designated plates ($|\mathcal{S}|\!=\!5$).}
    \label{fig:realworld_tasks}
\end{figure}

The evaluation comprises two long-horizon medical laboratory tasks with strong dependencies between successive steps, illustrated in Fig.~\ref{fig:realworld_tasks}.
\begin{itemize}
    \item \textit{Task 1: Blood Gas Analysis.}
    For each sample the robot must execute the seven atomic skills of Fig.~\ref{fig:realworld_tasks}: grasp the sample tube, remove its cap, discard the cap, dock with the analyzer, insert the tube onto the sampling probe, withdraw it once the assay finishes, and discard it. The approximately 5\,mm inlet clearance requires precise alignment, while the visually similar insertion and withdrawal stages create perceptual aliasing.

    \item \textit{Task 2: Blood Sample Tube Sorting.}
    Two categories of sample tubes (clotting and routine) are placed at randomized rack locations, and the robot must sort every tube into its designated plate. It demands less dexterity than Task~1 but more planning: the planner must infer each tube's position and category from the initial observation and commit to a feasible pick-and-place order, which System~2 tracks across many visually similar cycles so that no tube is skipped.
\end{itemize}

\subsubsection{Data Collection}
\label{subsec:data}
Expert operators collect all demonstrations by teleoperation, recording two complementary corpora per task. The first contains complete long-horizon demonstrations of the entire workflow, with each atomic skill's start and end boundaries annotated online to yield temporally aligned instruction--trajectory pairs. The second comprises single atomic-skill episodes from randomized initial poses, broadening state coverage and improving instruction-following robustness. We collect $1200$ episodes for \textit{Blood Gas Analysis} and $1000$ for \textit{Blood Sample Tube Sorting}.
Positive supervision for the recovery head comes from two further sources, both recorded during the same teleoperation sessions and kept out of the nominal corpora above. The first is deliberate intervention: an operator removes the sample from the gripper while the robot delivers it to the analyzer or places it during tube sorting, after which the run is terminated. The second is spontaneous teleoperation failure: when an operator mishandles a sample during collection, most often by dropping a tube on the way to the analyzer, collection stops immediately and the run is tagged as a failure episode rather than discarded. For each such episode the final $K_{\mathrm{rec}}$ frames receive positive labels $\beta_t=1$, with $K_{\mathrm{rec}}$ drawn uniformly from $[20,40]$ independently per episode so that the head is not tuned to one fixed offset from the terminating event. 

\begin{figure*}[!t]
    \centering
    \includegraphics[width=\textwidth]{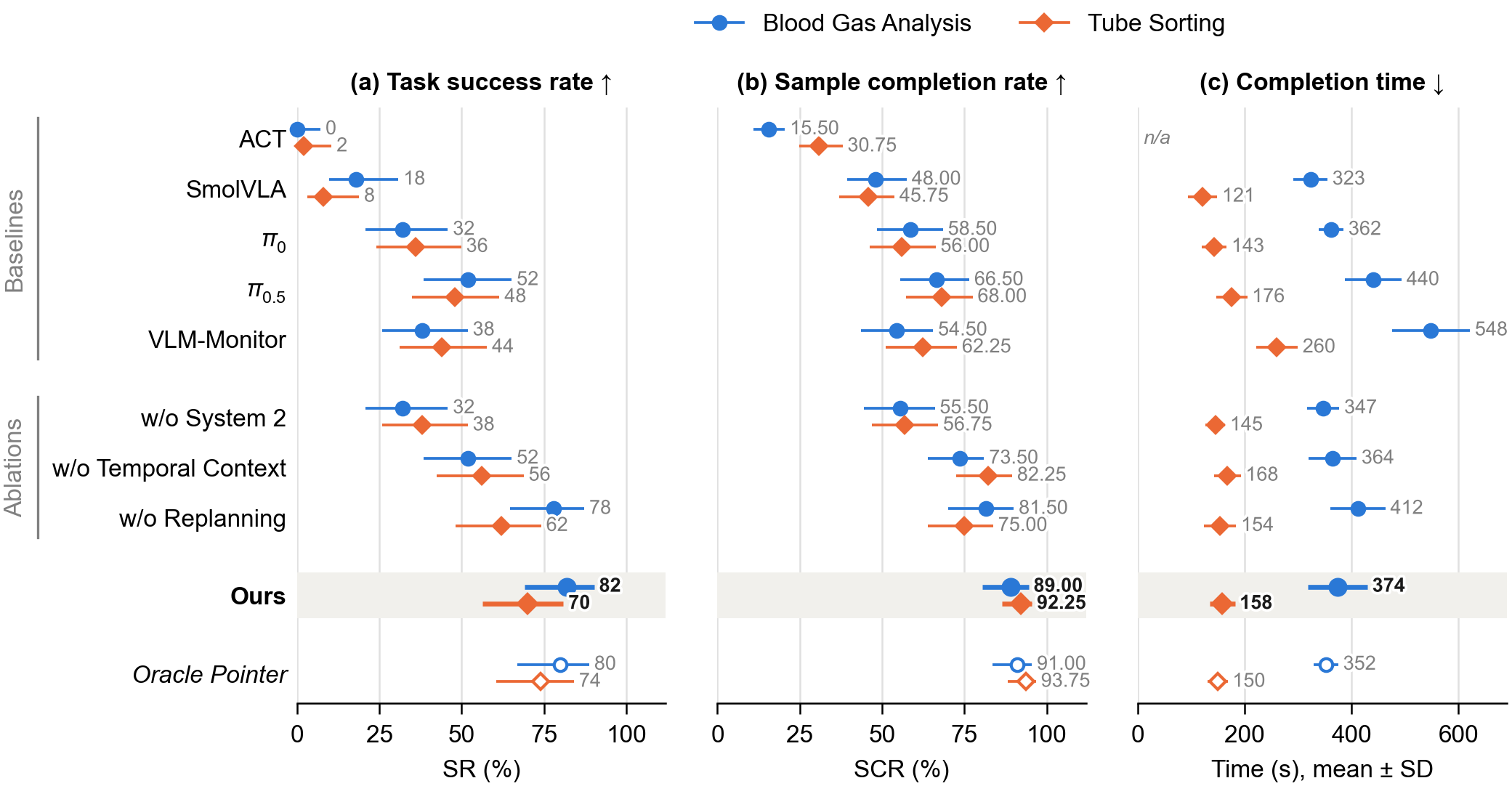}
    \caption{\textbf{Real-world performance on the nominal suite.} (a) Task success rate SR, (b) sample completion rate SCR, and (c) completion time. Error bars denote 95\% confidence intervals in (a)--(b) and $\pm1$ SD in (c).}
    \label{fig:main_results}
\end{figure*}

\subsubsection{Training Details}
The two systems are trained separately on the corpora of Sec.~\ref{subsec:data}, with the configurations listed in Table~\ref{tab:train_config}. For System~1 and every VLA baseline, a training sample pairs a multiview RGB observation and the concurrent proprioceptive state with the active atomic-skill instruction, and the supervised action horizon equals the deployed chunk length $N\!=\!30$. Starting from the released StarVLA weights~\cite{starvla2026codebase}, System~1 fine-tunes the Qwen3-VL backbone and the flow-matching action head jointly for each task. System~2 keeps the SigLIP2 encoders frozen and trains only the skill-grounding trunk and the two decision heads. The two terms of Eq.~\eqref{equ:eq6} carry equal weight, and $w_c$ and $w_r$ follow the inverse positive-class frequency of their labels so that the sparse positive frames are not swamped by negatives.

\begin{table}[!t]
    \centering
    \caption{Key runtime hyperparameters of the asynchronous dual-system controller.}
    \label{tab:hyperparams}
    \footnotesize
    \begin{tabular}{@{} c l c @{}}
        \toprule
        \textbf{Symbol} & \textbf{Meaning} & \textbf{Value} \\
        \midrule
        $f_{\mathrm{ctrl}}$ & control-loop rate & 20\,Hz \\
        $N$ & action-chunk length & 30 \\
        $r$ & chunk prefetch overlap & 8 \\
        $w$ & inter-chunk blend window (steps) & 4 \\
        $K$ & System-2 frame sampling interval & 15 \\
        $M$ & history-window length & 6 \\
        $\tau_{\mathrm{done}}$ / $k_a$ & advance threshold / debounce & 0.6 / 2 \\
        $\tau_{\mathrm{rec}}$ / $k_b$ & recover threshold / debounce & 0.8 / 3 \\
        $T_{\max}$ & per-skill stall timeout & 30\,s \\
        $\Delta_{\mathrm{cd}}$ & switch cooldown (steps) & 10 \\
        \bottomrule
    \end{tabular}
\end{table}

\begin{table}[!t]
    \centering
    \caption{Training configuration of System~1 and System~2.}
    \label{tab:train_config}
    \footnotesize
    \setlength{\tabcolsep}{4pt}
    \begin{tabular}{@{} l c c @{}}
        \toprule
        \textbf{Setting} & \textbf{System~1} & \textbf{System~2} \\
        \midrule
        Trainable modules   & backbone + action head & trunk + two heads \\
        Frozen modules      & --                     & SigLIP2 encoders \\
        Optimizer           & AdamW~\cite{loshchilov2019decoupled} & AdamW \\
        Learning rate       & $1\times10^{-4}$       & $1\times10^{-4}$ \\
        LR schedule         & cosine                 & cosine \\
        Batch size          & 8                      & 128 \\
        Training length     & 80\,k steps            & 60\,k steps  \\
        Loss weights        & --                     & $\lambda_c\!=\!\lambda_r\!=\!1$ \\
        \bottomrule
    \end{tabular}
\end{table}

\subsubsection{Baselines and Ablation Variants}
The baselines differ in how much of Robo-MLT they replace. Four policy baselines replace the entire dual-system pipeline with a single learned policy that must infer the active workflow stage itself: ACT~\cite{zhao2023learning} is an action-chunking imitation-learning policy, SmolVLA~\cite{shukor2025smolvla} and $\pi_0$~\cite{black2024pi_0} are generative VLA policies that predict continuous action trajectories, and $\pi_{0.5}$~\cite{intelligence2025pi_05} is a strong hierarchical baseline with subtask-level semantic reasoning, whose subtask supervision reuses the same atomic-skill boundary annotations that System~2 is trained on. All four are trained on the corpora of Sec.~\ref{subsec:data} with the same schedule as System~1, the VLA policies starting from their officially released checkpoints. \VarVLM{} instead retains the skill plan and pointer-based workflow tracking of System~2, while replacing the learned completion and recovery heads with zero-shot assessments from the Qwen3.7-Max API~\cite{yang2025qwen3}. Given the active-skill instruction and observations from the global and wrist cameras, the VLM determines whether the current skill has been completed and whether its execution remains feasible.

We additionally evaluate reduced variants of our own system. \VarNoSys{} removes System~2 together with the plan it tracks: a policy with the architecture and initialization of System~1, trained like the policy baselines, executes the entire workflow from the task-level instruction. The other variants retain the System~1 weights and the initial plan. \VarSingle{} ($M\!=\!1$) restricts the monitor to the current frame, removing temporal context while leaving both decision heads and the recovery pathway intact. \VarNoRec{} disables the high-level recovery pathway entirely: neither the recovery head nor the stall timeout can fire and no replanning is performed, so the plan pointer advances solely on completion-head decisions. \VarTimeout{} disables the recovery-head trigger but retains the stall timeout and the replanner, so recovery remains possible without the learned trigger. As a reference for switch timing, \VarOracle{} replaces System~2 with a human operator who advances the pointer by hand once the active skill is complete.

\subsubsection{Evaluation Protocol}
We evaluate each task under two conditions: nominal execution and sample removal, with 50 and 20 predefined configurations, respectively. Each configuration varies the initial tube poses and rack positions, while maintaining a fixed batch size of samples for blood gas analysis and tube sorting. All methods are evaluated on identical configurations within each condition, with interleaved execution orders and consistent robot initialization to ensure fair comparisons.

In the nominal condition, the robot executes the complete workflow without intervention, which tests end-to-end performance and the autonomous handling of execution failures. In the sample-removal condition, an operator removes the in-hand sample during execution, rendering the active skill infeasible, and the robot must revise its remaining workflow and process the available samples without further assistance. \VarNoRec{} is run under both conditions and \VarTimeout{} under sample removal only.

To isolate System~2 from System~1 execution errors, We evaluate the monitor offline using recorded demonstrations. For this evaluation, the monitors are retrained with the same architecture and objective using recordings that exclude the held-out workflows. Each workflow is replayed at the deployment cadence ($K$, $\tau_{\mathrm{done}}$, $k_a$, and $\Delta_{\mathrm{cd}}$ as in Table~\ref{tab:hyperparams}), with each skill boundary classified as on time, early, late, or missed. A workflow is considered correct only if all its skill boundaries are detected on time.

\subsubsection{Metrics}
We evaluate task performance at both workflow and sample levels. A sample is considered completed only when all required operations are executed in the prescribed order, including analysis and disposal for blood gas analysis or correct placement for tube sorting.

For nominal trials, success rate (SR) measures the percentage of trials in which all samples are successfully processed within the task-specific time limit. To capture partial progress, we additionally report the sample completion rate (SCR):
\begin{equation}
    \mathrm{SCR} = \frac{1}{|\mathcal{E}|}\sum_{e\in\mathcal{E}} \frac{n_e^{\mathrm{done}}}{n_{\mathrm{tot}}} \times 100\%
\end{equation}
where $\mathcal{E}$ denotes the evaluation trials, $n_e^{\mathrm{done}}$ is the number of completed samples in trial $e$, and $n_{\mathrm{tot}}$ is the total number of samples per trial.

For the sample-removal suite, we use the recovery success rate (RSR) to measure whether the robot can autonomously complete the remaining workflow after intervention:
\begin{equation}
    \mathrm{RSR}=\frac{1}{|\mathcal{E}_{\mathrm{rm}}|}\sum_{e\in \mathcal{E}_{\mathrm{rm}}}r_e \times 100\%
\end{equation}
where $\mathcal{E}_{\mathrm{rm}}$ denotes the disturbed trials, and $r_e=1$ if all remaining available samples are processed in protocol order within the execution limits and $r_e=0$ otherwise.

\subsection{Results}
After the end-to-end results, Secs.~\ref{subsec:offline}--\ref{subsec:runtime} examine the three requirements of Sec.~\ref{sec:intro} in turn, and Sec.~\ref{subsec:failure_analysis} locates the residual failures.

\begin{table}[!t]
    \centering
    \caption{Offline evaluation of System~2 on the blood-gas workflows.}
    \label{tab:offline}
    \footnotesize
    \setlength{\tabcolsep}{4pt}
    \begin{tabular}{@{} l c c c @{}}
        \toprule
        \textbf{Monitor} &
        \textbf{Workflows} &
        \textbf{On time / early /} &
        \textbf{Frame} \\
        & \textbf{all on time} &
        \textbf{late / missed} &
        \textbf{AP} \\
        \midrule
        \VarVLM{}
            & 10/50 & 96 / 220 / 14 / 5 & -- \\
        \midrule
        Ours, $M\!=\!1$
            & 18/50 & 172 / 148 / 12 / 3 & 0.767 \\
        Ours, $M\!=\!2$
            & 34/50 & 317 / 14 / 3 / 1 & 0.918 \\
        Ours, $M\!=\!4$
            & 39/50 & 321 / 5 / 4 / 5 & 0.941 \\
        \textbf{Ours}, $M\!=\!6$
            & \textbf{48/50} & 333 / 0 / 2 / 0 & 0.967 \\
        Ours, $M\!=\!8$
            & 32/50 & 309 / 0 / 26 / 0 & 0.942 \\
        \bottomrule
    \end{tabular}
\end{table}

\begin{table}[!t]
    \centering
    \caption{Recovery success after sample removal.}
    \label{recovery:tab:results}
    \footnotesize
    \setlength{\tabcolsep}{6pt}
    \renewcommand{\arraystretch}{1.05}
    \begin{tabularx}{0.85\columnwidth}{@{}
        >{\hsize=1.25\hsize\RaggedRight\arraybackslash}X
        >{\hsize=0.95\hsize\centering\arraybackslash}X
        >{\hsize=0.80\hsize\centering\arraybackslash}X @{}}
    \toprule
    \textbf{Method} & \textbf{Recovered/injected} & \textbf{RSR (\%)}\\
    \midrule
    \multicolumn{3}{c}{\textit{Blood Gas Analysis}}\\
    \VarOurs{}      & 18/20 & 90\\
    \VarTimeout{}   & 13/20 & 65\\
    \VarNoRec{}     & \phantom{0}0/20 & \phantom{0}0\\
    \midrule
    \multicolumn{3}{c}{\textit{Tube Sorting}}\\
    \VarOurs{}      & 17/20 & 85\\
    \VarTimeout{}   & 11/20 & 55\\
    \VarNoRec{}     & \phantom{0}0/20 & \phantom{0}0\\
    \bottomrule
    \end{tabularx}
\end{table}

\subsubsection{Workflow Completion}
\label{subsec:main_results}
\begin{figure*}[!t]
    \centering
    \includegraphics[width=\textwidth]{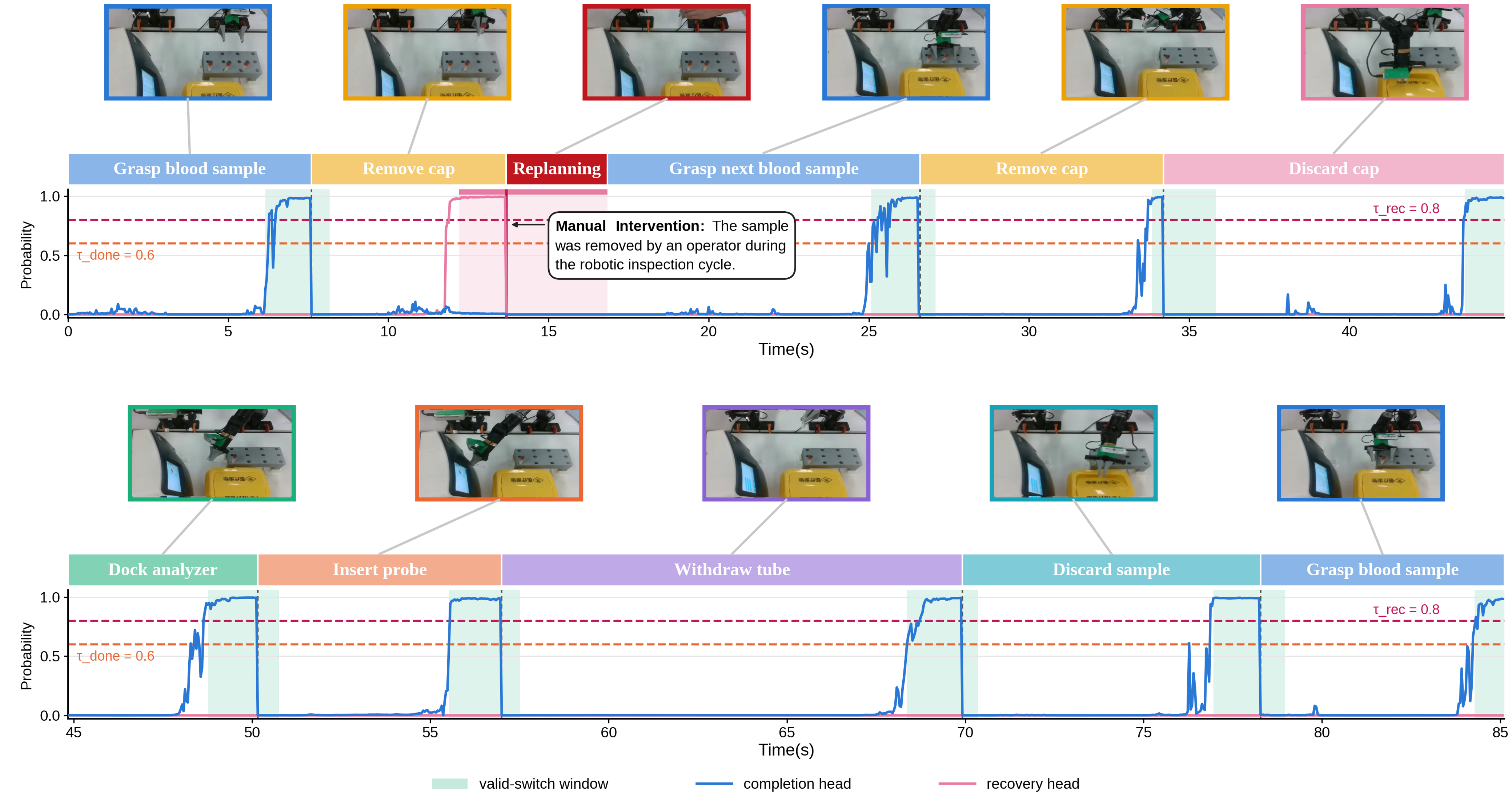}
    \caption{\textbf{Representative execution timeline of Robo-MLT on Blood Gas Analysis.} The completion and recovery probabilities predicted by System~2 are shown over a continuous execution, with the active skills indicated above each plot. The timeline illustrates skill transitions and workflow recovery following an operator-induced interruption.}
    \label{fig:experiment_sequences}
\end{figure*}

\begin{figure*}[!t]
    \centering
    \includegraphics[width=\textwidth]{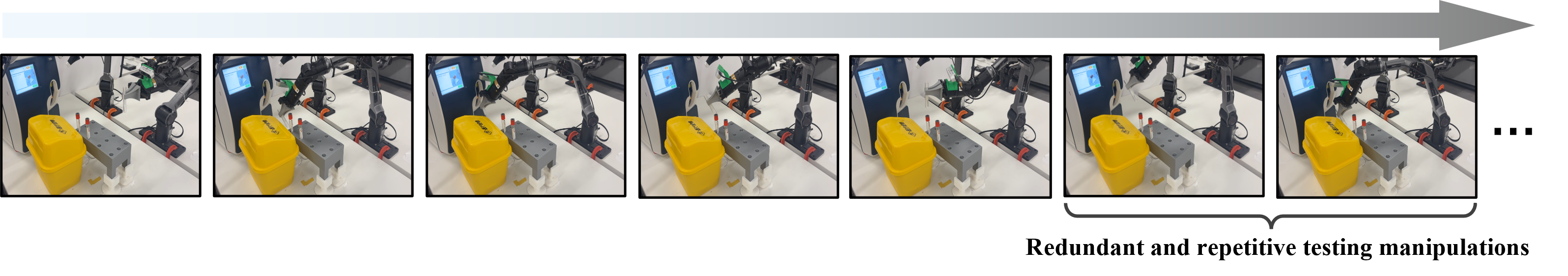}
    \caption{\textbf{Representative perceptual-aliasing failure of end-to-end IL and VLA baselines on Blood Gas Analysis.} Visually similar observations during insertion and withdrawal can cause the policy to confuse workflow stages, leading to repeated insertion attempts and ultimately stalling the procedure.}
    \label{fig:il_vla_failure}
\end{figure*}

Among autonomous methods, Robo-MLT achieves the highest SR and SCR on both tasks (Fig.~\ref{fig:main_results}); a representative run appears in Fig.~\ref{fig:experiment_sequences}. Its SR exceeds that of $\pi_{0.5}$ by 30 and 22 percentage points, respectively. This comparison is informative: $\pi_{0.5}$ is trained using the same skill-boundary annotations as System~2, but infers the active subtask at each step from current observations. In contrast, Robo-MLT maintains an explicit skill pointer and updates it using multiframe evidence. The perceptual-aliasing failure at the docking stage (Fig.~\ref{fig:il_vla_failure}) illustrates this distinction.

At the sample level, Robo-MLT achieves SCRs of 89.00\% and 92.25\%, outperforming $\pi_{0.5}$ by 22.5 and 24.25 percentage points. Across 50 configurations, these gains correspond to 45 additional fully processed blood gas samples and 97 additional correctly sorted tubes. Robo-MLT also achieves shorter mean completion times on both tasks, suggesting that the additional monitoring does not introduce an execution-time penalty.

Removing System~2 isolates the performance of the executor without workflow-level supervision. When trained to execute the entire workflow directly from a task-level instruction, \VarNoSys{} achieves SRs of only 32\% and 38\%, with SCRs of 55.50\% and 56.75\%. These results are comparable to those of $\pi_0$ and remain below $\pi_{0.5}$. The full framework improves SR by 50 and 32 percentage points, respectively. The improvement over policy baselines thus appears to stem primarily from workflow management rather than differences in System~1 architecture.

\VarVLM{} evaluates whether a general-purpose VLM can replace the learned execution monitor in closed-loop operation. Although it uses the same plan, skill pointer, and System~1 as Robo-MLT, it achieves SRs of only 38\% and 44\%, underperforming $\pi_{0.5}$ on both tasks. It also exhibits the longest and most variable completion times (548.4 $\pm$ 73.2\,s and 260.3 $\pm$ 39.0\,s), consistent with the overhead of API-dependent decisions. Its inferior performance relative to \VarSingle{} mirrors the offline replay results (Sec.~\ref{subsec:offline}), indicating that zero-shot VLM inference cannot reliably replace a task-specific learned monitor. By contrast, Robo-MLT performs skill-switching decisions locally on the GPU and reserves API calls for planning and replanning, thereby decoupling online monitoring from network latency.

\begin{table}[!t]
    \centering
    \caption{Runtime and computational overhead on the deployment GPU.}
    \label{tab:runtime}
    \footnotesize
    \setlength{\tabcolsep}{6pt}
    \renewcommand{\arraystretch}{1.05}
    \begin{tabularx}{0.85\columnwidth}{@{} >{\RaggedRight\arraybackslash}X r @{}}
        \toprule
        \textbf{Quantity} & \textbf{Value} \\
        \midrule
        System~1 chunk inference latency        & 271\,ms \\
        System~2 monitor inference latency      & 259\,ms \\
        End-to-end control delay                & 364\,ms \\
        Sustained control rate                  & 20\,Hz ($\pm0.5$) \\
        System~2 update rate                    & 1.33\,Hz \\
        Peak GPU memory (System~1 $+$ System~2) & 10.14\,GB \\
        \bottomrule
    \end{tabularx}
\end{table}

\VarOracle{} serves as a human-guided reference for subtask transitions while using the same System~1 policy. It obtains SRs of 80\% and 74\% and SCRs of 91.00\% and 93.75\% on blood gas analysis and tube sorting, respectively. Robo-MLT approaches this reference without human intervention: its SR differs by only one and two trials out of 50 episodes, respectively. Notably, human-guided switching still leaves 10 and 13 of the 50 trials unsuccessful. This suggests that residual errors are largely attributable to within-skill System~1 execution, rather than solely to the accuracy of subtask switching.

Replacing the multiframe monitor with its single-frame counterpart (\VarSingle{}, $M\!=\!1$) reduces SR from 82\% to 52\% on blood gas analysis and from 70\% to 56\% on tube sorting. As this variant preserves the plan, skill pointer, prediction heads, and System~1, the degradation isolates the contribution of temporal context. In particular, blood gas SR drops to that of $\pi_{0.5}$ (52\%), highlighting the role of temporal evidence in the full framework's performance advantage. A single-frame monitor may trigger premature transitions, such as initiating cap removal before a sample has been grasped, thereby violating the prerequisites of subsequent operations (Sec.~\ref{subsec:offline}).

Disabling workflow-level recovery (\VarNoRec{}) decreases SR by 4 and 8 percentage points, but SCR by 7.5 and 17.25 percentage points, corresponding to 15 fewer processed blood gas samples and 69 fewer correctly sorted tubes. This discrepancy reflects the difference between trial-level and sample-level metrics: losing one sample already causes a trial to fail under SR, whereas recovery determines whether the remaining samples can still be processed. Without recovery, the robot may persist at an infeasible skill until the time limit. With recovery enabled, Robo-MLT revises the remaining workflow and continues processing the other samples, preventing a single failure from blocking the entire batch.

\subsubsection{Offline Analysis of the Execution Monitor}
\label{subsec:offline}
We next isolate workflow-stage identification, the first requirement described in Sec.~\ref{sec:intro}. In closed-loop execution,failures may arise from either mistimed skill transitions or System~1 execution errors. Offline  recorded demonstrations eliminates the latter source, allowing the results in Table~\ref{tab:offline} to be attributed to the monitor. With $M\!=\!6$, the monitor correctly identifies nearly all skill boundaries without premature transitions, and the vast majority of workflows exhibit no mistimed switches. These results suggest that transition timing is not the primary performance bottleneck, consistent with Robo-MLT's performance approaching \VarOracle{} in closed-loop evaluation  (Sec.~\ref{subsec:main_results}).

The window ablation explains the value of observation history. A skill nearing completion can appear nearly identical to one that has just completed in a single frame. Consequently, short windows produce premature transitions. Conversely, overly long windows introduce detection delays: immediately after completion, most observations still depict the preceding motion, yielding 26 late transitions for $M\!=\!8$. These errors have different implications. Premature switching can initiate an operation before its prerequisites are satisfied, as illustrated by early cap removal (Sec.~\ref{subsec:main_results}), whereas delayed switching primarily increases execution time. Among the evaluated settings, $M\!=\!6$ is the shortest window that eliminates premature transitions. Accordingly, workflow-level switching accuracy in offline replay offers a practical criterion for selecting the window length when adapting the monitor to new protocols, without requiring additional robot rollouts. 

\subsubsection{Recovery After Sample Removal}
\label{subsec:disturbance}
Sample removal deprives the active skill of its target, so System~1 cannot complete it by adapting its motion. \VarNoRec{} confirms this: it recovers in none of the 40 trials (Table~\ref{recovery:tab:results}), so every recovery in the table comes from the workflow-level pathway, as in the rollout of Fig.~\ref{fig:experiment_sequences}. With that pathway, Robo-MLT reaches RSRs of 90\% and 85\%, against 65\% and 55\% for \VarTimeout{}. The two variants share the same replanner and differ only in when replanning starts. The recovery head needs $k_b\!=\!3$ consecutive updates above $\tau_{\mathrm{rec}}$, about 1.5\,s of sustained evidence, whereas the timeout fires only once the interrupted skill has lasted $T_{\max}\!=\!30$\,s. During that wait, System~1 keeps executing an instruction whose target is gone, outside the coverage of its demonstrations, which may leave the workspace in a state that the revised plan cannot handle. 

\subsubsection{Runtime and Computational Overhead}
\label{subsec:runtime}
In Table~\ref{tab:runtime}, the end-to-end control delay is the cost of one synchronous cycle: 364\,ms to refresh the multiview observation and robot state and to generate an action chunk from them, of which 271\,ms is System~1 inference. Operated synchronously, the arm would therefore halt for 364\,ms after every 1.5\,s chunk ($N/f_{\mathrm{ctrl}}$), idling about a fifth of the time and lengthening every motion by about a quarter. The asynchronous runtime takes that cost off the control path by overlapping chunk generation with execution, and the 259\,ms monitor finishes well within the 0.75\,s between its windows ($K/f_{\mathrm{ctrl}}$), so the controller holds its nominal 20\,Hz rate. With a peak of 10.14\,GB for both systems, the full stack runs on a single consumer GPU beside the bench rather than on a dedicated server.

\subsubsection{Failure Analysis}
\label{subsec:failure_analysis}
The residual failures of Robo-MLT concentrate in two within-skill execution modes of System~1 that occur while the active skill is correct. In blood gas analysis the dominant mode is docking misalignment: the grasped tube is not centered on the analyzer's probe needle and fails to seat in the ${\sim}5$\,mm-clearance inlet, so insertion remains incomplete. In tube sorting it is a placement error: System~1 releases the tube into a plate other than the one the active instruction designates. Because the workflow state remains nominal in both modes, neither head is designed to intercept them. Improving System~1 at these two steps is therefore the most direct route to higher workflow success.

\subsection{Limitations}
Robo-MLT has several limitations. System~1 and System~2 are learned from the same demonstrations but trained in separate stages, which adds engineering complexity. The skill library is extensible only in a restricted sense: the frozen planner re-orders and repeats existing skills freely, but a new laboratory operation requires demonstrations for System~1 and retraining of System~2, and the recovery head covers only the failure families seen in training. The sample-removal disturbance used in Sec.~\ref{subsec:disturbance} belongs to one of those families, so the reported RSR measures recovery from an in-distribution interruption and does not establish generalization to unseen ones. The two-task evaluation also uses the same workspace and object distributions as data collection, so robustness to broader scene shifts remains untested. Future work will relax the closed skill library toward open-ended instruction following, and strengthen System~1 through reinforcement fine-tuning and knowledge-isolation training that keeps action data from eroding the semantic priors of the VLM backbone.

\section{Conclusion}
We presented Robo-MLT, a hierarchical dual-system framework for long-horizon manipulation in medical laboratory workflows. A frozen vision--language planner orders atomic skills, System~2 tracks the active skill from multiframe observations and invokes replanning of the remaining steps once that skill can no longer be completed, and System~1 generates continuous actions with a VLM-conditioned flow-matching policy under an asynchronous runtime.
On both laboratory tasks, Robo-MLT completes more workflows and carries more samples through the full workflow than the policy baselines and the VLM-based monitor, leaving fewer batches for a technologist to finish by hand. It also resumes the remaining workflow after an operator removes the sample being handled, which the same system without workflow-level recovery never does, and it runs at the robot control rate on a single GPU. These results indicate that explicitly representing workflow progress can improve long-horizon execution when different task stages produce similar observations. Recovery is evaluated on a single disturbance type in a single workspace; extending the skill library and the disturbance coverage to a broader set of laboratory protocols remains future work.

\bibliographystyle{IEEEtran}
\bibliography{ref/ref_RMLT}
\end{document}